\pdfoutput=1

\documentclass[11pt]{article}

\usepackage[final]{acl}

\usepackage{times}
\usepackage{latexsym}

\usepackage[T1]{fontenc}

\usepackage[utf8]{inputenc}

\usepackage{microtype}

\usepackage{inconsolata}

\usepackage{graphicx}

\usepackage{booktabs}
\usepackage{multirow}

\usepackage{makecell}
\usepackage{bbding}
\usepackage{multirow}
\usepackage{pgfplots}
\usepackage{svg}
\usepackage{amsmath} 
\usepackage{amssymb}
\usepackage{url}
\usepackage{tablefootnote}
\usepackage{graphicx}
\usepackage{cuted}
\usepackage{tabularx}
\usepackage{xcolor}
\usepackage{pifont}

\usepackage{adjustbox}
\usepackage[utf8]{inputenc} 
\usepackage[T1]{fontenc}    

\usepackage{graphicx}   
\usepackage{stmaryrd} 
\usepackage{cancel}     
\usepackage{amssymb}    

\usepackage{graphicx}   
\usepackage{tikz}       
\usepackage{amssymb}    

\usepackage{ulem} 
\usepackage{tipa} 
\DeclareUnicodeCharacter{0259}{\textschwa} 

\usepackage{listings}
\newcommand{\benchmarkname}{DocMEdit}

\newcommand{\halfcheck}{
    \CheckmarkBold\hspace{-0.375cm}\XSolidBrush
}
\newcommand{\halfchecktext}{
    \CheckmarkBold\hspace{-0.355cm}\XSolidBrush
}

\usepackage[most]{tcolorbox}
\definecolor{deepgreen}{RGB}{6,153,6} 
\definecolor{deepred}{RGB}{254,34,35}  
\definecolor{DarkYellow1}{RGB}{188, 158, 0}

\title{\benchmarkname: Towards Document-Level Model Editing}

\author{
 \textbf{Li Zeng\textsuperscript{1\footnotemark[1]}},
 \textbf{Zeming Liu\textsuperscript{2\footnotemark[1]}},
 \textbf{Chong Feng\textsuperscript{1}},
 \textbf{Heyan Huang\textsuperscript{1}},
 \textbf{Yuhang Guo\textsuperscript{1\footnotemark[2]}}
\\
 \textsuperscript{1}School of Computer Science and Technology, Beijing Institute of Technology, Beijing, China\\
 \textsuperscript{2}School of Computer Science and Engineering, Beihang University, Beijing, China
\\
\href{zengli@bit.edu.cn}{\textcolor{black}{\{zengli}},
\href{fengchong@bit.edu.cn}{\textcolor{black}{fengchong}},
\href{hhy63@bit.edu.cn}{\textcolor{black}{hhy63}},
\href{guoyuhang@bit.edu.cn}{\textcolor{black}{guoyuhang\}}}@bit.edu.cn
\href{zmliu@buaa.edu.cn}{\textcolor{black}{zmliu}}@buaa.edu.cn
}

\begin{document}
\maketitle
\begin{abstract}
\renewcommand{\thefootnote}{\fnsymbol{footnote}} 
\footnotetext[1]{Equal contribution}
\footnotetext[2]{Corresponding author: \href{guoyuhang@bit.edu.cn}{guoyuhang@bit.edu.cn}}
\renewcommand{\thefootnote}{\arabic{footnote}}

Model editing aims to correct errors and outdated knowledge in the Large language models (LLMs) with minimal cost.
Prior research has proposed a variety of datasets to assess the effectiveness of these model editing methods.  
However, most existing datasets only require models to output short phrases or sentences, overlooks the widespread existence of document-level tasks in the real world, raising doubts about their practical usability.
Aimed at addressing this limitation and promoting the application of model editing in real-world scenarios, we propose the task of document-level model editing. To tackle such challenges and enhance model capabilities in practical settings, we introduce \benchmarkname, a dataset focused on document-level model editing, characterized by document-level inputs and outputs, extrapolative, and multiple facts within a single edit.
We propose a series of evaluation metrics and experiments.
The results show that the difficulties in document-level model editing pose challenges for existing model editing methods\footnote{Dataset and codes are publicly available at \url{https://github.com/BITHLP/DocMEdit}}.

\end{abstract}
\begin{figure}[!ht]
    \centering    \includegraphics[width=\linewidth,height=1.45\linewidth]{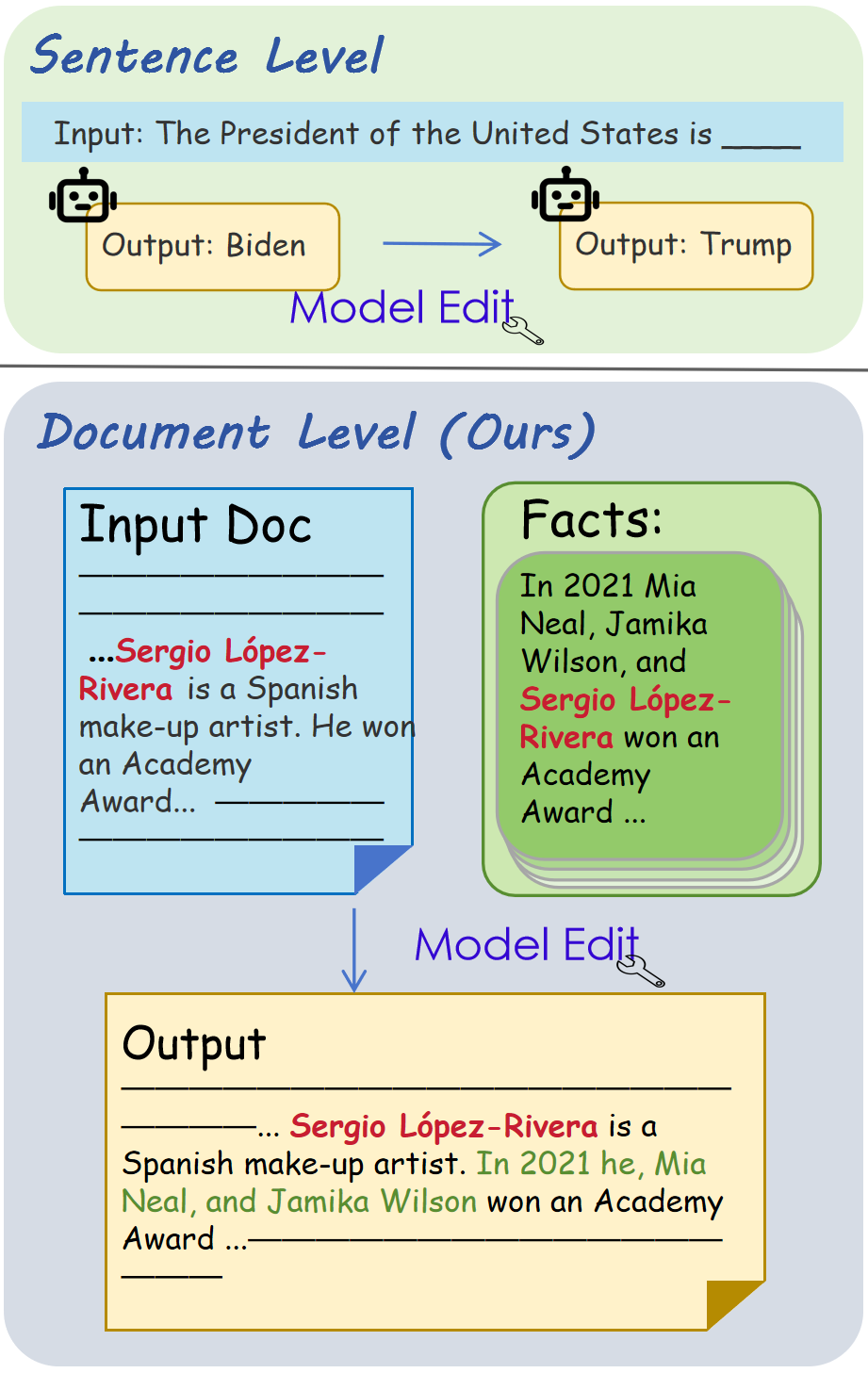}
    \caption{\label{figure:example}An example of \benchmarkname.  The input and output of \benchmarkname~are both document-level contents. Model editing should inject multiple facts to be edited into the model, enabling the edited model to output the updated document.
    }
\end{figure} 

\section{Introduction}
Large language models (LLMs) have demonstrated exceptional performance across a wide range of fields and are widely applied in various practical scenarios \citep{touvron2023Llama,touvron2023Llama2,wang2024survey,geva2020transformer,geva2022transformer}. Given their broad usage, it is crucial for LLMs to deliver accurate and reliable information. However, LLMs may still generate incorrect or outdated information due to the knowledge they store, which can be inaccurate \citep{decaoetal2021editing, agarwal2022temporal}. Such inaccuracies can lead to serious consequences in critical domains, such as medical diagnoses and legal advice, highlighting the importance of methods to correct errors in LLMs. To address this issue without expensive retraining, model editing techniques have been proposed \citep{mitchell2022memory,Sinitsin2020Editable,decaoetal2021editing}.

To evaluate the effectiveness of model editing methods, previous researchers have proposed a range of datasets encompassing various tasks such as single-hop and multi-hop question answering, cloze tasks, and others \citep{zeng2024fame,xie2024memlaenhancingmultilingualknowledge,gu-etal-2024-model}. To verify whether a model knows specific facts, almost all of these datasets require the model to output a short phrase or sentence. 
As shown in figure \ref{figure:example}, prior datasets only demand shorter textual outputs from the model. However, in real-world scenarios, document-level tasks such as generating biographies, long chains of thought, or updating Wikipedia documents are more common.
Moreover, the document-level model editing challenge for model editing methods lies in the need to extrapolate answers from facts, handle longer contexts, and deal with multiple facts within a single document.
Due to the lack of research on these challenges, the application of existing model editing methods to practical LLMs is limited.

To promote the application of model editing in document-level tasks, we propose a new task: \textbf{document-level model editing}.
Document-level model editing requires both the input and output to be at the document level. Additionally, the model cannot infer the answer from the given facts; it involves longer contexts and multiple facts within a single edit.

Aimed at addressing the lack of practical applicability in existing benchmarks and promoting model editing at the document level, we propose a novel benchmark: \benchmarkname, a model editing dataset that contains document-level data. 
The input and output of \benchmarkname~are both at the document level. Besides, the model cannot derive the answer solely from the facts to be edited; instead, it must combine multiple facts to be edited with existing knowledge to produce the updated document. Unlike most previous research, our editing facts are not derived from triples but are directly extracted from unstructured data in Wikipedia, which is more aligned with real-world model editing scenarios.
\benchmarkname~ contains 37,990 data items, with an average context length of 1,535.5 per item. It includes 105,652 editing facts, with an average of 2.78 facts per data item. Additionally, we extracted triples from \benchmarkname~that share the same relations as those in Wikidata to facilitate experiments with RAG-based methods \citep{zeng2024fame,zhang2024oneeditneuralsymboliccollaborativelyknowledge,zhang-etal-2024-knowledge-graph}.

To validate the effectiveness of existing model editing methods for document level model editing tasks, we propose a series of new metrics and conduct experiments on \benchmarkname. We also further discuss how each challenge in document level model editing affects the performance of model editing methods. The conclusion demonstrates that existing methods have low accuracy while exhibiting strong side effects. Further research confirms that factors such as document length, fact length, number of facts, and fact updates all impact the effectiveness of model editing methods.

The main contributions of this paper are as follows:
    \begin{itemize}
    \setlength{\itemsep}{0em}
    \setlength{\parskip}{0pt}
    \setlength{\parsep}{0pt}
        \item  We are the first to propose the document-level task for model editing, which is more aligned with real-world LLM use situations.

         \item To support research in document-level model editing, we create \benchmarkname, a novel benchmark that includes longer contexts and multiple parallel edit facts within a single document.

    \item Experiments show that existing methods have low accuracy and significant side effects. Further research demonstrates that the length of the document and facts, the number of facts, and the fact updates all impact the performance of model editing methods.
    \end{itemize}

\section{Related work}

    \subsection{Model Editing Datasets}

    Model editing datasets serve the purpose of verifying the effectiveness of methods and enhancing the capability of LLMs. However, existing datasets focus on question answering with shorter contexts, neglecting the common scenario of long context input-output pairs in practice, i.e., document-level model editing. Tasks in existing datasets include QA, sentence completion, choose, and cloze tests \citep{zeng2024fame,zhong2023MQuAKE,levy2017zero,meng2022locating,wang2023knowledge,10.1145/3698590}. \citet{wu2024akew} explored model editing for longer texts, however, their expected output is still at the sentence level, and no supporting facts are provided. Unlike previous research, our input and output are both at the document level. Additionally, we provide multiple facts to be edited and require the LLMs to generate an updated document based on these facts.
    
    \subsection{Document-Level NLP}

    Currently, some existing researches discuss the challenges that arise when common NLP tasks are scaled to the document level, such as translation \citep{wang-etal-2023-document-level}, relation extraction \citep{xue2024autore,zheng2024comprehensive}, and QA \citep{rasool2024evaluating}. These researches highlight the lack of suitable datasets, evaluation methods, and the limitations of existing models in accurately information retrieve. However, they all overlook the field of model editing. In contrast, we are the first to scale model editing to the document level, introducing a corresponding dataset, experiments, and evaluation metrics.

\begin{table*}[!ht]  
 \setlength{\tabcolsep}{4pt}
\centering
\renewcommand{\arraystretch}{0.5}
\scalebox{0.74}[0.72]{%
 \renewcommand{\arraystretch}{0.8}
\begin{tabular}{cccccccc}
\toprule
\textbf{Benchmark} & \textbf{Doc level} & \textbf{Extrapolative} & \textbf{Multi-Edits} & \textbf{Locality} & \textbf{Avg. Target Len.} & \textbf{Avg. Facts Len.} & \textbf{Total} \\[-2pt]
\midrule
\textbf{\textsc{ZsRE} \citep{levy-etal-2017-zero}} 
& \color{deepred}\XSolidBrush & \color{deepred}\XSolidBrush & \color{deepred}\XSolidBrush &\color{deepgreen}\CheckmarkBold  & 12.12 & 56.39 & 270.0K \\[-2pt]
\textbf{\textsc{CounterFact} \citep{meng2022locating}} 
& \color{deepred}\XSolidBrush & \color{deepred}\XSolidBrush & \color{deepred}\XSolidBrush &\color{deepgreen}\CheckmarkBold  & 6.65 & 39.14 & 2.2K \\[-2pt]
\textbf{\textsc{MQuAKE} \citep{zhong2023MQuAKE}} 
& \color{deepred}\XSolidBrush & \color{deepred}\XSolidBrush & \color{deepgreen}\CheckmarkBold & \color{deepred}\XSolidBrush & 10.94 & 145.09 & 11.1K \\[-2pt]
\textbf{\textsc{WikiBio} \citep{manakul-etal-2023-selfcheckgpt}} 
& \color{DarkYellow1}{\halfcheck} & \color{deepred}\XSolidBrush & \color{deepred}\XSolidBrush & \color{deepred}\XSolidBrush & 131.10* & 131.10* & 1.4k \\
\textbf{\textsc{FAME} \citep{zeng2024fame}} 
& \color{deepred}\XSolidBrush & \color{deepred}\XSolidBrush & \color{deepgreen}\CheckmarkBold & \color{deepgreen}\CheckmarkBold & 13.01 & 62.75 & 128.0K \\[-1pt]
\midrule
\textbf{\textsc{\benchmarkname} (Ours)} 
& \color{deepgreen}\CheckmarkBold & \color{deepgreen}\CheckmarkBold & \color{deepgreen}\CheckmarkBold & \color{deepgreen}\CheckmarkBold & 867.62 & 623.40 & 38.0K \\[-2pt]
\bottomrule
\end{tabular}
}
\caption{Comparison of benchmarks."Doc level" refers to whether the input and output of the dataset are at the document level. \color{DarkYellow1}{\halfchecktext} \color{black} means the input is at the document level, but the output is at the sentence level. "Extrapolative" refers to whether the answer to each question requires inference based on the existing knowledge within the LLM, or if it can be directly derived from the given facts alone. "Multi-Edits" refers to whether multiple facts to be edited are included in a single editing target. "Locality" refers to whether the dataset design takes into account the evaluation of side effects. "Avg. Target Len." refers to the average expected output length per data item, and "Avg. Facts Len." denotes the average total length of the facts used per data item. "Total" represents the total amount of data. "*" indicates that the expected output is identical to the facts to be edited.}
\label{tab:benchmark_comparison}
\end{table*}
    
\section{Problem Definition}
Model editing aims to modify the knowledge contained within a model, changing the output related to the facts to be edited while keeping other outputs unchanged. Based on previous research \citep{zeng2024fame,wang2023knowledge,yao2023editing}, we define document level model editing as follows:



The document-level model editing task aims to modify the document-level output of a large language model based on a set of facts to be edited while keeping the parts unrelated to these facts unaffected. Specifically, let the large language model be denoted as $\mathcal{M}$, with the input and output of the model represented as \( x \) and \( y \), respectively, i.e., \( y = \mathcal{M}(x) \). The output \( y \) consists of multiple sentences, represented as $y = \sum_{i=1}^n s_i$, where $s_i$ are the sentences of the original document. In the document level model editing, the original large language model $\mathcal{M}$ generates an output \( y \) based on the input document \( x \), and then the model is edited using a set of facts to be edited $ F = \sum_{j=1}^m f_j$, resulting in the edited model $\mathcal{M'}$, which generates a new output document \( y' = \mathcal{M'}(x) \).

The new document \( y' \) consists of the original sentences as well as newly added sentences supported by the facts to be edited. It is represented as $y' = \sum_{i=1}^n s_i + \sum_{j=1}^k s_{f_j}$, where $s_i$ are the sentences consistent with the original text, and $s_{fj}$are the new sentences supported by the facts to be edited. In this process, it is required that all sentences unrelated to the facts to be edited, $s_i$, remain unchanged, while the newly added sentences,$s_{fj}$, must be content supported by the facts $f_j$.

\label{sec:problemsDefinition}
\section{\benchmarkname: A Dataset Dedicated to Document Level Model Editing}

\begin{figure}[!ht]
    \centering
    \includegraphics[width=\linewidth,height=\linewidth]{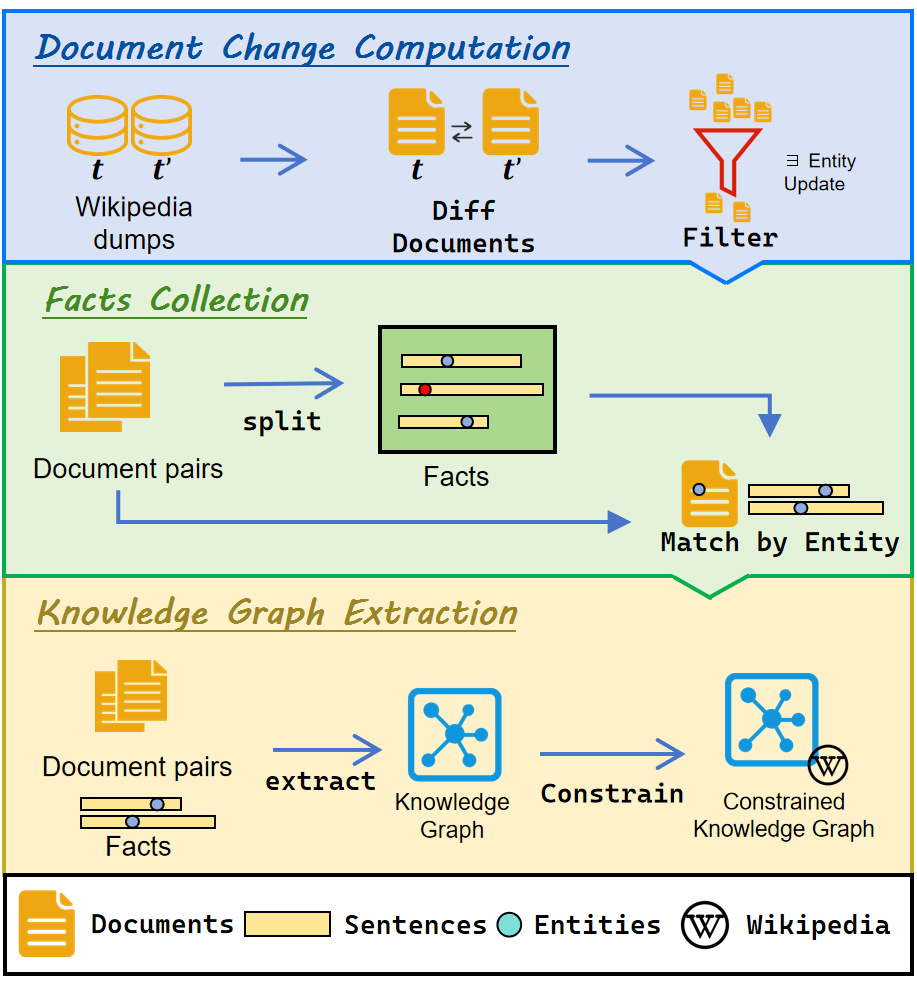}
    \caption{\label{figure:dataset}The construction process of \benchmarkname. In the Document Change Computation, we calculate the updates of documents in the Wikipedia between two time points and retain those documents that exhibit entity updates. In the Facts Collection, based on the newly added entities within the documents, we identify the newly added sentences mentioning these entities and extract them as supporting facts. In the Knowledge Graph Extraction, we extract structured knowledge graphs and impose constraints based on Wikidata relations.
    }
\end{figure} 


To advance the practical application of document-level model editing, we introduce \benchmarkname~(\textbf{Doc}ument level \textbf{M}odel \textbf{Edit}). The input and output in \benchmarkname~are document-level, with each document containing one or more facts to be edited. We also extract triples from each document and fact to facilitate the use of knowledge graph-based model editing methods. 

\subsection{Dataset Construction}  
Following \citet{logan2021fruit}, we construct \benchmarkname~with the following steps: (1) Document Change Computation; (2) Facts Collection; (3) Knowledge Graph Extraction. The detailed steps are as follows:
\subsubsection{Document Change Computation}  
We collect Wikipedia dumps from two timestamps (20231101 and 20241101). 
For each document in the dump, we extract its INTRODUCTION section, using the corresponding sections from two timestamps as \( y \) and \( y' \), respectively. We consider this as reflecting the updates in Wikipedia.
Since most Wikipedia updates are stylistic rather than factual \citep{daxenberger-gurevych-2012-corpus}, we filtered out updates that did not include the addition of at least one entity to ensure that the extracted updates were meaningful.  

\subsubsection{Facts Collection}  
We extract sentences related to entities from the document as facts. Specifically, following the assumption of \citet{logan2021fruit}, for each sentence, if an entity mentioned in the sentence was newly introduced in the document update, the sentence was considered to support the document update \( f_i \) of the corresponding entity. 

\subsubsection{Knowledge Graph Extraction}  
Given that many model editing methods leverage knowledge graphs relevant to facts for better performance \cite{zeng2024fame,zhang-etal-2024-knowledge-graph}, we utilize the framework proposed by \citet{schmitz2012open} to extract knowledge graphs. We extract knowledge graphs from the source document, the target document, and the supporting facts, respectively. Then we constrain the relation \(r\) in the triple \((s, r, o)\) to be an existing relation in Wikidata, which ensures that our extracted knowledge graphs are well-structured and consistent.

\begin{table}[!ht]
\centering
\renewcommand{\arraystretch}{0.9}
\begin{tabular}{c|ll} 

\toprule
\multirow{4}{*}{\benchmarkname}  & Data Items          & 37990   \\
                                 & Avg. Context Len. & 1535.5  \\
                                 & Facts             & 105651  \\
                                 & Avg. Facts        & 2.78    \\
\hline
\multirow{3}{*}{\makecell{knowledge \\ graph}} & Entities          & 568652  \\
                                 & Relations         & 4804    \\
                                 & Triples           & 1411057\\
\bottomrule
\end{tabular}
\caption{\label{tab:benchmark_statistics}Dataset statistics. "Data Items" refers to the total number of data entries, and "Facts" denotes the facts to be edited. We also collected statistical information on the extracted triples, which were derived from the input documents, output texts, and editing facts. Note that the context length includes both the input and the expected output, while not including the facts.}
\end{table}

\subsection{Quality Control}

Following \citet{logan2021fruit}, we adopt the following measures to ensure the quality of our dataset: (1) manually removing unsupported updates, and (2) comparing the annotated data with automatically collected text to verify the reliability of the data collection process. We evaluate our dataset using the metrics described in Section~\ref{sec:metrics}, yielding a DR of 81.17 and a DE of 89.71, which demonstrate a high consistency between our dataset and human annotations, confirming that facts in our dataset are aligned with the documents and that the document updates are well-supported.

\subsection{Benchmark Analysis}
\subsubsection{Comparison}

\benchmarkname~is distinguished by its document-level input and output. Additionally, it is uniquely characterized by the feature that the expected output cannot be directly derived from the facts to be edited. \benchmarkname~includes cases where a document corresponds to multiple facts, while also allowing for the testing of side effects in model editing methods. Finally, \benchmarkname~has an advantage in terms of context length, which has been overlooked by previous datasets.

\subsubsection{Statistics}
Table \ref{tab:benchmark_statistics} presents the statistics of \benchmarkname. We have compiled statistics on the documents, facts, and the extracted knowledge graph.
For data samples and additional statistics, please refer to Appendix \ref{app:dataset_sample}.
\section{Experiment}
\label{sec:main_exp}
In this section, we present our main experiments.
We introduce the edited models used in section \ref{sec:exp_llm}, the baseline model editing methods in section \ref{sec:exp_baseline}, and the evaluation metrics in section \ref{sec:metrics}. We analyze the results of our experiments in section \ref{sec:sub_main_res}. For more experimental details, please refer to Appendix \ref{app:exp_detail}.

\subsection{Language Models}
\label{sec:exp_llm}
Following \citet{zeng2024fame}, we use LLMs of various sizes to evaluate the performance of the model editing method in handling diverse scenarios. Since we aim to test the update of internal knowledge in LLMs, we want the model to be unaware of the updated facts during training so we intentionally choose earlier released LLMs. These models include GPT-2 XL \citep{solaiman2019release}, GPT-J-6B \citep{wang2021gpt},  Llama2 \citep{touvron2023Llama2} and Deepseek-V3 \cite{liu2024deepseek}. Since the EREN method is specifically designed for instruction models, we also design experiments on  Mistral-7B-Instruct-v0.3  \citep{jiang2023mistral7b}.

\subsection{Baselines}

\label{sec:exp_baseline}
Following \citet{zhang2024comprehensivestudyknowledgeediting}, we selected the following methods as baselines for evaluation. For parameter modification methods, we choose FT and MEMIT, and for parameter preservation approaches, we select IKE, SKEME, and EREN as baselines.
Please refer to Appendix \ref{app:Implementation details} for the implementation details.

\paragraph{FT} The most classic and straightforward model-editing method is fine-tuning. Following previous research \citep{meng2022mass}, we apply Fine-Tuning to the given layer of the model.

\paragraph{MEMIT \citep{meng2022mass}} Currently considered a state-of-the-art method among parameter modification methods.

\paragraph{IKE \citep{zheng2023can}} Direct embedding similarity-based RAG method.

\paragraph{SKEME \citep{zeng2024fame}} RAG method based on knowledge graphs and caching systems, which allows for more precise knowledge retrieval.

\paragraph{EREN \citep{chen2024robust}} RAG method based on a growing notebook, capable of handling updates to multiple facts.

\subsection{Metrics}

\label{sec:metrics}
Following the experimental setups of previous studies \citep{meng2022mass,decaoetal2021editing,yao2023editing}, this work evaluates both the effectiveness and side effects of model editing. In addition, we adopt the evaluation protocols from \citet{meng2022locating} and \citet{zeng2024fame} to assess the generation quality of the edited models as well as the editing efficiency of different model editing methods.

\paragraph{Accuracy}
Accuracy is used to evaluate the effectiveness of editing. We measure accuracy on two levels: at the level of each editing target (i.e., each document update) and the level of each piece of fact. The former evaluates the overall outcome of the update, while the latter assesses whether the model has appropriately utilized each piece of fact.

For both levels, we employ two metrics to evaluate editing performance: ROUGE and an entity-based measurement. For ROUGE, we use UpdateROUGE following \citet{logan2021fruit}, which only computes the score on the parts of the source and target documents that were actually updated. For entity-based evaluation, we extracted all entities involved in the documents or sentences to assess the ability of methods to update facts. These two levels and two metrics result in four evaluation methods: Document-ROUGE (\textbf{DR}), Document-Entity (\textbf{DE}), Edit-ROUGE (\textbf{ER}), and Edit-Entity (\textbf{EE}).  Please refer to Appendix \ref{app:Metrics format def} for the formal definitions of thest metrics.

\paragraph{Locality}
Model editing requires that edits do not affect outputs unrelated to the edited facts. In document-level model editing, outputs unrelated to the edited facts correspond to the unchanged parts of the document before and after the update. We evaluate the side effects of model editing methods by computing both ROUGE and entity-based differences on the unchanged parts of the document, denoted as ROUGE Side Effect (\textbf{RSE}) and Entity Side Effect (\textbf{ESE}). Please refer to Appendix \ref{app:Metrics format def} for the formal definitions of thest metrics.

\paragraph{Quality}
\citet{meng2022locating} recommend evaluating the impact of model editing on the generation quality of large language models. Following \citet{9699426,liu-etal-2020-towards-conversational}, we adopt human evaluation to evaluate the semantic coherence (\textbf{SC}) of model outputs. Specifically, the outputs of large language models are categorized into three quality tiers, with detailed scoring criteria provided in Appendix~\ref{app:sc}. Note that since this metric relies on human evaluation, we assess a sample of 100 data points for each method.

\paragraph{Efficiency}

To evaluate whether model editing methods can perform edits efficiently, we follow the approach of previous researches \cite{zeng2024fame, yao2023editing} by measuring  time consumption (\textbf{Ti}) and memory requirements (\textbf{Me}).

\begin{table*}[!ht]
    \renewcommand{\arraystretch}{0.85}
\begin{tabular*}{\hsize}{@{}@{\extracolsep{\fill}}cccccccccccc@{}}
      \hline

        \multirow{2}*{Model} & \multirow{2}*{Method}&\multicolumn{4}{c}{Accuracy}  &  \multicolumn{2}{c}{Locality}   &  Quality & \multicolumn{2}{c}{Efficiency}   \\
       \cmidrule(lr){3-6} \cmidrule(lr){7-8}  \cmidrule(lr){9-9} \cmidrule(lr){10-11}
        ~&  ~&    DR$\uparrow$&
        DE$\uparrow$ & ER$\uparrow$ & EE$\uparrow$ &  RR$\uparrow$&ER $\uparrow$ &SC $\uparrow$ &Ti $\downarrow$ &Me $\downarrow$  \\
     
        \hline
        
        \multirow{5}*{GPT2-XL} 
        & w/o Edit &\underline{13.66} & \underline{24.12} & \underline{13.71} & 2.25 & \textbf{43.34}& \textbf{31.95} &1.01 &  9.35 & 8.98 \\
         ~ & FT &   {12.08 }& 11.72 & 11.55 & 3.00 & 37.36 & \underline{18.07}  & 0.54&  \underline{10.16} & 12.84  \\
        ~ & MEMIT&   11.86 & 9.64 & 12.05 & 3.00 &\underline{37.41} & 13.88  & 0.64& \textbf{10.06 }& 12.84  \\
        ~ & IKE &  \textbf{14.10} & \textbf{31.77} & \textbf{14.93} & \textbf{9.25} & 34.38 & 17.82  & \underline{1.02}& 10.58 & \underline{10.02}   \\
        ~ & SKEME&   10.71 & 15.59 & 11.98 & \underline{5.50} & 36.11 & 12.55  & \textbf{1.43}&  10.62 & \textbf{9.98} \\
        
       \hline

        \multirow{5}*{GPT-J} & 
         w/o Edit & \textbf{22.08} & 16.42 & \underline{17.26} & 5.50 & \textbf{53.37 }& \textbf{52.86 } &\underline{1.05} & 40.43 & 24.62 \\
         ~ & FT &  2.44 & \textbf{35.00} & 4.90 & 1.00 & 33.38 & 1.53&0.48 &  \underline{67.56} & 31.49  \\
        ~ & MEMIT&   9.79 & 22.82 & 10.17 & 5.50 & 33.99 & 6.10  &0.52 & 72.94 & 32.58 \\
        ~ & IKE &   17.05 & \underline{29.68}& 16.81 & \underline{11.58} & 40.01 & 26.22& 0.98&  42.56 & \textbf{27.62} \\
        ~ & SKEME&   \underline{19.53} & 28.60 & \textbf{25.77} & \textbf{23.83} & \underline{42.05} & \underline{38.33}&\textbf{1.08} & \textbf{40.57}& \underline{28.22}  \\
      
       \hline

        \multirow{5}*{Llama2}
        &  w/o Edit &  \textbf{26.11} & 18.97 & 15.77 & 0.50 & \textbf{53.91}& \textbf{55.37}&\textbf{1.05}&30.84 & 27.64  \\
         ~ & FT &    \underline{24.78}& 17.95 & 14.65 & 7.17 & \underline{53.76}& 39.22&0.60&  47.51 & \underline{31.49}\\
        ~ & MEMIT&   19.63 & 9.62 & 15.16 & 2.50 & 40.54 & 34.86 &0.62&64.42 & 32.58  \\
        ~ & IKE &  19.79 & \underline{26.30} & \underline{22.77} & \underline{12.20} & 43.27 & 35.80  &\underline{1.03}&\textbf{31.70} & 33.67 \\
        ~ & SKEME&      21.08 & \textbf{29.34} & \textbf{25.75} & \textbf{23.92} & 47.31 & \underline{49.22} &1.00&\underline{32.28} & \textbf{30.92} \\

       \hline

        \multirow{4}*{Mistral}
        &  w/o Edit &   3.62 & 4.56 & 1.95 & 1.25 & 33.39 & 8.59& \underline{1.94}& 23.92 & 28.34   \\
        ~ & IKE &    5.29 & 24.02 & 5.52 & 10.00 & 33.58 & 11.79  & \textbf{1.95}&\textbf{26.17} & \underline{29.37}  \\
        ~ & EREN &    \textbf{11.26} & \underline{27.94}&\underline{12.70}& \textbf{18.77} & \underline{35.63} & \textbf{25.60}  &1.92 & \underline{26.18} & 30.31 \\
        ~ & SKEME&   \underline{10.11} & \textbf{35.03}&\textbf{12.94}& \underline{11.75} & \textbf{36.27}& \underline{24.17} & \textbf{1.95}&27.64 & \textbf{29.26} \\

       \hline

        \multirow{4}*{DeepSeek}
          & w/o Edit & 34.81 & 23.62 & 19.08 & 27.22 & 54.97 & 88.80 &\textbf{1.99}  &- & - \\
    ~ & IKE & \underline{37.43} & \textbf{37.68} & 25.41 & 45.45 & \textbf{63.76} & \textbf{92.35} &\underline{1.93}&- & -  \\
    ~ & EREN & 36.91 & 23.05 & \underline{29.48} & \textbf{55.90} & \underline{61.29} & \underline{90.16} & 1.91& -& - \\
    ~ & SKEME &  \textbf{37.71}  &  \underline{37.05} & \textbf{29.64} & \underline{54.49} & 59.04 & 88.55 & \textbf{1.99} & -& - \\
        
      \hline
    \end{tabular*} 
    \caption{    \label{tab:mainresult}Main result on \benchmarkname. "DR", "DE", "ER", "EE", "RSE", "ESE", "SC", "Ti", and "Me" stand for Document-ROUGE, Document-Entity, Edit-ROUGE, Edit-Entity, ROUGE Side Effect, Entity Side Effect, semantic coherence, time consumption, and memory requirements, respectively. "w/o Edit" represents the unedited original model. Bold indicates the best-performing method, while underlining denotes the second-best. In the evaluation of editing efficiency, comparisons are made only among methods that involve model editing. "-": Editing efficiency is unavailable for DeepSeek, as it is evaluated via API calls.
}

\end{table*}

\subsection{Main Result}
\label{sec:sub_main_res}
Table \ref{tab:mainresult} shows the results of our main experiments. It can be observed that all the base models fail to handle the document-level model editing. We have the following conclusions.

\textbf{The overall performance of the methods was below expectations.} For each base model, when they have not been edited, they achieved a relatively high DR, indicating that the generated documents are quite similar to the target edits. However, since no editing facts were provided to the base models, both DE and EE are low, which suggests that a high DR may be attributed to hallucinations, whereas DE and EE better reflect fact-based modifications.

We found that methods involving parameter modifications have lower ER and EE compared to RAG-based methods, implying that they have a lower ability to edit facts in the document-level model editing task setup. The methods for modifying parameters all cause a certain degree of degradation in the generation quality of LLMs, which may be due to partially impairing the generative capability of these models.

For RAG-based methods, EREN does not achieve satisfactory results. We found that the core objective of EREN is to determine whether a given fact is relevant to the input. However, in our dataset, every provided fact is related to the document being edited. The challenge for the model is to decide which facts should be utilized and where they should be incorporated, which differs from the relevance problem tackled by EREN. For SKEME and IKE, since SKEME relies on knowledge graph search while IKE uses vector knowledge base search, and fact retrieval based on documents is more challenging, SKEME has a significant advantage (see Appendix \ref{app:retrval_result} for comparisons).


\textbf{All models suffer from significant side effects.} We found that each model exhibits serious side effects. For the base model, the ESE of all LLMs is below 60, meaning that they lose more than 40\% of the entity information. After editing, all models show a significant decrease in both RES and ESE, indicating that these methods have a substantial impact on parts of the document unrelated to the edited facts.

\section{Analysis}
\label{sec:analysis}
In this section, we conduct a deeper analysis to examine the key differences between document-level model editing and traditional model editing, as well as the challenges posed by these differences.

We contend that the main differences between document-level model editing and traditional model editing are: 1) the length of the input and facts to be edited is longer, and 2) multiple parallel facts to be edited are allowed within a single editing objective. 

To substantiate our conclusions, we design a series of research questions (RQs) and analyze their impact on editing effectiveness. For the first aspect, we propose the following research questions:  
\textbf{RQ1a}: The impact of context length on editing.  
\textbf{RQ1b}: The impact of fact length on editing.
For the second aspect, we propose:  
\textbf{RQ2a}: The impact of the number of edited facts per document on the results. 
\textbf{RQ2b}: The impact of fact updates on model editing performance. 

According to the findings of the main experiments, entity-based metrics are more indicative of fact-based modifications. Therefore, in the analysis, we use DE to evaluate document-level RQ1a and RQ2b, and adopt EE to assess edit-level RQ2a. 
For all RQs, we conduct experiments on Llama2.

\subsection{\textbf{RQ1a}: The Impact of Context Length on Editing. }
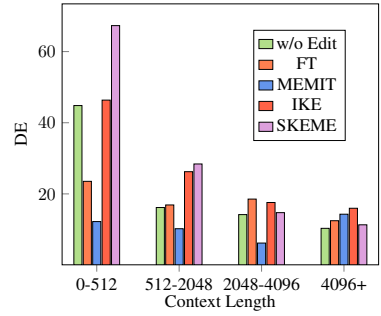
\begin{figure}[!htbp]
\centering
\resizebox{140pt}{!}{%
\begin{tikzpicture} 
\begin{axis}[
    ybar=2*\pgflinewidth,
    symbolic x coords={0-512, 512-2048, 2048-4096, 4096+},
    xtick=data,
    ylabel={DE},
    ylabel style={at={(0.05,0.5)}},
    xlabel={Context Length},
    legend style={at={(0.75, 0.9)}, anchor=north, legend columns=1},
    enlarge x limits={abs=0.75cm},
    bar width=5pt,  
    xtick distance=1,  
    legend image code/.code={\draw[#1,fill=#1] (0cm,-0.1cm) rectangle (0.3cm,0.1cm);}, 
    xtick pos=bottom,    
    ytick pos=left,      
    axis lines=box,      
]
\definecolor{color1}{RGB}{178,223,138} 
\definecolor{color2}{RGB}{255,127,80}  
\definecolor{color3}{RGB}{100,149,237} 
\definecolor{color4}{RGB}{255,99,71}   
\definecolor{color5}{RGB}{221,160,221} 

 \addplot[fill=color1] coordinates {(0-512, 44.83) (512-2048, 16.17) (2048-4096, 14.18) (4096+, 10.32) };
 \addplot[fill=color2] coordinates {(0-512, 23.53) (512-2048, 16.89) (2048-4096, 18.54) (4096+, 12.46) };
 \addplot[fill=color3] coordinates {(0-512, 12.22) (512-2048, 10.21) (2048-4096, 6.18) (4096+, 14.29) };
 \addplot[fill=color4] coordinates {(0-512, 46.36) (512-2048, 26.24) (2048-4096, 17.58) (4096+, 15.96) };
 \addplot[fill=color5] coordinates {(0-512, 67.3) (512-2048, 28.41) (2048-4096, 14.73) (4096+, 11.31) };
\legend{w/o Edit, FT, MEMIT, IKE, SKEME}
\end{axis}
\end{tikzpicture}
}
\caption{\label{fig:rq1a}result of RQ1a. The x-axis represents the context length of the data, while the y-axis represents the corresponding DE.}
\end{figure}

To analyze the impact of context length on the model's output, we categorized the data based on context length and then evaluated the performance of each method within each context length range, the results are shown in Figure \ref{fig:rq1a}.

Figure \ref{fig:rq1a} presents the results of RQ1. For shorter documents (length 0-512), the base model can already handle some cases, while IKE and SKEME further improve performance. However, FT and MEMIT actually degrade the model’s performance. For longer documents (length 512-1024), only RAG-based learning methods (IKE and SKEME) are effective. For even longer documents, almost all methods fail to generate successfully edited documents.

\subsection{\textbf{RQ1b}: The Impact of Fact Length on Editing. }
To analyze the impact of fact length on editing performance, we measure the EE of each fact and arrange them according to the length of the facts.

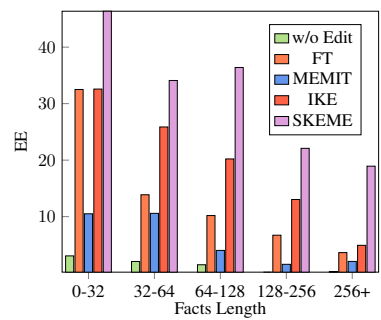
\begin{figure}[!htbp]
\centering
\resizebox{140pt}{!}{%
\begin{tikzpicture}
\begin{axis}[
    ybar=2*\pgflinewidth,
    symbolic x coords={0-32, 32-64, 64-128, 128-256, 256+},
    xtick=data,
    ylabel={EE},
    ylabel style={at={(0.05,0.5)}},
    xlabel={Facts Length},
 legend style={at={(0.8, 0.95)}, anchor=north, legend columns=1},
    enlarge x limits={abs=0.75cm},
    clip=false,
    enlargelimits=true,
    ytick={0, 10, 20, 30, 40, 50, 55}, 
    enlarge y limits=false, 
    axis on top=false,
    bar width=5pt,  
    xtick distance=1,  
    legend image code/.code={\draw[#1,fill=#1] (0cm,-0.1cm) rectangle (0.3cm,0.1cm);}, 
    xtick pos=bottom,    
    ytick pos=left,      
    axis lines=box,      
]
\definecolor{color1}{RGB}{178,223,138} 
\definecolor{color2}{RGB}{255,127,80}  
\definecolor{color3}{RGB}{100,149,237} 
\definecolor{color4}{RGB}{255,99,71}   
\definecolor{color5}{RGB}{221,160,221} 

 \addplot[fill=color1] coordinates {(0-32, 3.05) (32-64, 2.06) (64-128, 1.48) (128-256, 0.17) (256+, 0.26) };
 \addplot[fill=color2]coordinates {(0-32, 32.5) (32-64, 13.87) (64-128, 10.19) (128-256, 6.71) (256+, 3.63) };
 \addplot[fill=color3] coordinates {(0-32, 10.5) (32-64, 10.58) (64-128, 4.02) (128-256, 1.56) (256+, 2.07) };
 \addplot[fill=color4] coordinates {(0-32, 32.58) (32-64, 25.87) (64-128, 20.21) (128-256, 13.03) (256+, 4.92) };
 \addplot[fill=color5] coordinates {(0-32, 46.36) (32-64, 34.1) (64-128, 36.4) (128-256, 22.09) (256+, 18.93) };
\legend{w/o Edit, FT, MEMIT, IKE, SKEME}
\end{axis}
\end{tikzpicture}
}
\caption{\label{fig:rq1b}result of RQ1b. The x-axis represents the length of the facts to be edited, while the y-axis represents the corresponding EE.}
\end{figure}

Figure \ref{fig:rq1b} presents the results of RQ1b. The findings indicate that as the length of the facts increases, the editing performance of all methods declines. Additionally, we observe that when the facts are relatively short, FT can achieve a certain level of editing effectiveness, whereas for longer facts, RAG-based methods become a better choice.  

Further analysis of the two RAG-based methods reveals that as fact length increases, the performance of IKE drops rapidly, while SKEME experiences a more gradual decline. We attribute this to the increased difficulty of retrieving longer facts using vector-based retrieval in IKE, while entity-based retrieval in SKEME remains relatively robust. Therefore, although both IKE and SKEME are affected by the LLM’s ability to utilize knowledge, IKE also suffers from the negative impact of fact length on its retrieval capability.

\subsection{\textbf{RQ2a}: The Impact of the Number of Edited Facts per Document on the Results.  }

\begin{figure}[!htbp]
\centering
\resizebox{140pt}{!}{%
\begin{tikzpicture}
\begin{axis}[
    ybar=2*\pgflinewidth,
    symbolic x coords={1, 2, 3, 4, 5+},
    xtick=data,
    ylabel={DE},
    xlabel={Number of Facts},
    ylabel shift=-10pt, 
    ylabel style={at={(0.05,0.5)}},
 legend style={at={(0.8, 0.9)}, anchor=north, legend columns=1},
    enlarge x limits={abs=0.75cm},
    bar width=5pt,  
    xtick distance=1,  
    legend image code/.code={\draw[#1,fill=#1] (0cm,-0.1cm) rectangle (0.3cm,0.1cm);}, 
    xtick pos=bottom,    
    ytick pos=left,      
    axis lines=box,      
]
\definecolor{color1}{RGB}{178,223,138} 
\definecolor{color2}{RGB}{255,127,80}  
\definecolor{color3}{RGB}{100,149,237} 
\definecolor{color4}{RGB}{255,99,71}   
\definecolor{color5}{RGB}{221,160,221} 

 \addplot[fill=color1] coordinates {(1, 24.0) (2, 15.43) (3, 12.58) (4, 18.28) (5+, 13.24) };
 \addplot[fill=color2] coordinates {(1, 21.23) (2, 12.93) (3, 12.48) (4, 15.93) (5+, 17.98) };
 \addplot [fill=color3]coordinates {(1, 14.2) (2, 4.57) (3, 4.33) (4, 5.42) (5+, 7.01) };
 \addplot [fill=color4]coordinates {(1, 33.05) (2, 22.12) (3, 28.99) (4, 17.35) (5+, 14.97) };
 \addplot[fill=color5] coordinates {(1, 40.2) (2, 24.02) (3, 25.93) (4, 17.83) (5+, 12.55) };
\legend{w/o Edit, FT, MEMIT, IKE, SKEME}
\end{axis}
\end{tikzpicture}
}
\caption{\label{fig:rq2a}result of RQ2a. The x-axis represents the number of edits corresponding to each document, while the y-axis represents its DE.}
\end{figure}
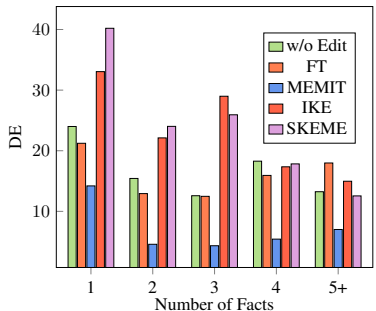

We track the number of editing facts involved in each document, and figure \ref{fig:rq2a} shows our experimental results. We found that the performance of all methods declines as the number of editing facts increases. When the number of editing facts is small, RAG-based methods perform better. However, as the number of editing facts grows, their performance rapidly deteriorates. When the number of editing facts reaches 5 or more, the FT method outperforms the RAG-based methods.

\subsection{\textbf{RQ2b}: The Impact of Fact Updates on Model Editing Performance.}

To test the impact of fact updates on document-level model editing, we followed the same process as in \benchmarkname. We collected updates from Wikipedia between 20220420 and 20231101 and merged them with the updates in \benchmarkname~ based on the document. As a result, we obtained a series of document updates between 20220420 and 20241101, along with their corresponding facts to be edited. Since these facts to be edited were derived from two separate updates, there are contradictions between them. We used this data construction method to simulate real-world fact updates in the model.

\begin{table}[!ht]
    \renewcommand{\arraystretch}{0.9}
    \centering
    \resizebox{\linewidth}{!}{%
    \begin{tabular}{ccccc}  

      \hline
        \multirow{2}*{Method}&\multicolumn{4}{c}{Accuracy}    \\
		\cline{2-5}
        ~&    DR$\uparrow$&
        DE$\uparrow$ & ER$\uparrow$ & EE$\uparrow$   \\
     
        \hline

w/o Edit & \textbf{24.86}\textsubscript{-1.25} & 17.83\textsubscript{-1.14} & 16.83\textsubscript{+1.06} & 0.47\textsubscript{-0.03} \\
FT & \underline{22.61}\textsubscript{-2.17} & 13.19\textsubscript{-4.76} & 9.75\textsubscript{-4.90} & 6.22\textsubscript{-0.95} \\
MEMIT & 16.14\textsubscript{-3.49} & 8.70\textsubscript{-0.92} & 12.15\textsubscript{-3.01} & 2.48\textsubscript{-0.02} \\
IKE & 18.52\textsubscript{-1.27} & \underline{22.13}\textsubscript{-4.17} & \underline{19.88}\textsubscript{-2.89} & \underline{10.83}\textsubscript{-1.37}\\
SKEME & 19.97\textsubscript{-1.11} & \textbf{27.45}\textsubscript{-1.89} & \textbf{22.16}\textsubscript{-3.59} & \textbf{19.59}\textsubscript{-4.33} \\

      \hline
    \end{tabular} }
    \caption{\label{tab:rq2b}Result of RQ2b. The numbers in the subscript represent the difference between the results and the main experiment results.
} 
\end{table}
\begin{table*}[!ht]
\centering
\fontsize{9}{10}\selectfont 
    \renewcommand{\arraystretch}{0.95}

\begin{tabular}{p{2.2cm}|p{5.6cm}p{5.6cm}|c}

\hline
Error Type  & Document and Facts   & Target and Generated & Ratio \\ \hline
Hallucination & Document: Ian David George (1953–2016) was...
\vspace{0.5cm}

Facts: None & 
\textcolor{deepred}{Ian David George (1953–2016) \uline{(also Tata Ian George or simply Tata)} was...}

\textcolor{deepgreen}{Ian David George (1953–2016) was...} &78.4  \\
\hline
Unexpected Style Change & Document: The middle class refers to a class of people in the middle of a social hierarchy, often defined by occupation, income, education, or social status...
\vspace{0.5cm}

Facts: None & 
\textcolor{deepred}{
The middle class \uline{is a social class that traditionally is} defined by occupational, income, educational, or social status level that is neither high nor low in a hierarchy....}

\textcolor{deepgreen}{
 The middle class refers to a class of people in the middle of a social hierarchy, often defined by occupation, income, education, or social status.} & 7.7\\ 
\hline
Ignoring Fact Update & Document:
Ashaghy Gushchular is a village in the Shusha District of Azerbaijan...
\vspace{0.5cm}

Facts: 'Yuxari Quscular: It was part of Shusha District with Malibeyli and Ashaghy Gushchular villages till 5 December 2023.' & 

\textcolor{deepred}{
Ashaghy Gushchular is a village in the Shusha District of Azerbaijan...}

\textcolor{deepgreen}{
Ashaghy Gushchular is a village in the Shusha District of Azerbaijan... \uline{It was part of the Shusha District with Malibeyli and Yuxari Quscular villages till 5 December 2023.}}  & 8.6\\


\hline
Misunderstanding  Facts& Document: Aremark is a municipality in Viken county, Norway...
\vspace{0.5cm}

Facts: Østfold have 17 (former 18) municipalities: \# Aremark...
& 
\textcolor{deepred}{
\uline{Østfold} is a county and former municipality in Norway.}

\textcolor{deepgreen}{
Aremark is a municipality in Østfold county, Norway.} & 5.3\\
\hline

\end{tabular}%
\caption{The four main error types that occur in \benchmarkname~ are Hallucination, Unexpected Style Change, Ignoring Fact Update, and Misunderstanding Facts. \textcolor{deepred}{Red} represents the generated output, \textcolor{deepgreen}{green} represents the target output. The \uline{underscored tilde} represents the missing/incorrect/extraneous parts. Some of the output has been truncated for clearer presentation. We only selected facts relevant to the displayed input. "None" indicates that there are no relevant supporting facts for the corresponding text. "Ratio" refers to the proportion of this error type among all erroneous outputs.}
\label{tab:error_analysis}
\end{table*}

Table \ref{tab:rq2b} presents the experimental results. The analysis shows that almost all methods experienced a decline in performance when dealing with fact updates. Specifically, since the input document was switched from the 20231101 version to the corresponding 20220420 document, the output quality of the unedited model also decreased. FT and MEMIT showed significant declines both at the document level and at the edit level. This decline is attributed to excessive continuous editing, which caused the internal parameters of the LLM to diverge too far from the initial state, leading to a drop in output quality \citep{gupta2024rebuildingromeresolving}. 

IKE and SKEME experience a noticeable decrease in edit-level accuracy. We speculate this is due to the fact conflicts and the overwhelming number of facts affect the context learning ability. Although SKEME was shown in its original paper to handle fact updates \citep{zeng2024fame}, the experiments reveal that it struggles with unstructured, document-level fact updates as seen in \benchmarkname.

\subsection{Discussion of Analysis}
In the aforementioned RQs, we confirmed that both longer inputs and facts influence the effectiveness of model editing. Moreover, multiple parallel facts and structured facts also degrade the performance of existing model editing methods. To address this issue, we argue that decomposing the overall editing task \citep{zhong2023MQuAKE,fei2024retrieval}, adjusting the prompt structure and the positioning of facts \citep{liu2024lost}, simultaneously attending to shallow and deep neurons as well as attention heads \citep{zhang2024locate}, and managing conflicts between internal and external knowledge within the model \citep{zhao2024steering} are all beneficial strategies.

\section{Error Analysis}
\label{sec:error_analysis}
We conducted an error analysis to identify the challenges in document level model editing. 

\subsection{Error categorization}
As shown in Table \ref{tab:error_analysis}, there are four main types of errors: Hallucination, Unexpected Style Change, Ignoring Fact Update, and Misunderstanding Facts.

\paragraph{Hallucination} The model incorrectly adds content to the document without factual support.

\paragraph{Unexpected Style Change} The model incorrectly alters the narrative style of the document without factual support. Unlike hallucination, an Unexpected Style Change does not involve changes to facts or entities, but merely modifies the style of narration. While this may seem harmless, the model still makes changes beyond the intended editing facts, which contradicts the definition provided in section \ref{sec:problemsDefinition}.

\paragraph{Ignoring Fact Update} Despite the provided editing facts containing the necessary information for document updates, the model fails to make the corresponding changes.

\paragraph{Misunderstanding Facts} The model successfully matches the facts with the document and attempts to make the necessary changes. However, it fails to produce the correct result and instead erroneously alters other parts of the document.

\subsection{Detailed Analysis and Discussion}
We manually annotated 100 output samples for each model-method combination to identify their corresponding error types. Table~\ref{tab:error_analysis} presents the distribution of several common error categories. Among them, hallucination errors are the most frequent, which is consistent with the findings of our main experiments.

For potential solutions, we recommend using the following approaches to address the four types of errors. For Hallucinations, \citet{wang2025gradient} suggest focusing on attention weights during editing rather than the commonly used FFN. For Unexpected Style Change, the key is to ensure that the model modifies facts while remaining faithful to the original text \cite{yao2025reff}. For Ignoring Fact Update, adjusting the placement of relevant facts can help the model focus on crucial contextual information \cite{liu2024lost,parasaram2024factselectionproblemllmbased}. For Misunderstanding Facts, enhancing the LLM’s comprehension ability, particularly in long-context scenarios, is essential \cite{an2024make}.

\section{Conclusion}
We introduce document level model editing, a model editing task that is more aligned with real-world applications. To advance research in this task, we present \benchmarkname, a dataset focused on document-level model editing. Experiments conducted on this dataset show that existing methods struggle with document-level model editing. Further experiments indicate that the challenges of document level model editing stem from long contexts and the presence of multiple facts to edit within a single document, aspects that are overlooked by current methods. We hope that our research will propel the field of model editing forward and inspire further research in this area.
\section*{Limitation}
Since \benchmarkname~focuses on document-level model editing, the inputs and outputs are relatively long, encompass numerous facts. This imposes substantial demands on the context length supported by LLMs and requires greater computational resources for processing.
\section*{Ethical Statement}
This study uses publicly available documents from Wikipedia and complies with its licensing requirements. The data involved in this study only includes publicly disclosed information. The study adheres to ethical guidelines, ensuring that the use of Wikipedia data is solely for academic and non-commercial purposes, and strives to mitigate any potential biases or unfair statements in the data. All ethical considerations regarding the privacy and potential harms associated with data usage have been carefully addressed.

\section*{Acknowledgments}
We thank all the anonymous reviewers for their insightful and valuable comments. This work is supported by the National Natural Science Foundation of China (Grant No. U21B2009, 62406015) and Beijing Institute of Technology Science and Technology Innovation Plan (Grant No. 23CX13027).

\normalem
\bibliography{custom}

\appendix

\section{Terminology Explanation}
In this section, we will explain some of the terms we used and their calculation methods.

\paragraph{Target Len}
Target Len is used to describe the content expected to be generated by the LLM. For question-answering or sentence completion datasets (Zsre, CounterFact, MQUaKE, FAME), it refers to the expected generated phrases. For Wikibio, it refers to the content generated after the original document. For \benchmarkname, it refers to the updated document.

\paragraph{Facts Len}
Facts Len refers to the length of the facts to be edited. For question answering or sentence completion datasets, it refers to the concatenation of all questions and answers. For Wikibio, it refers to the content expected to be generated by the model. For \benchmarkname, it refers to the facts to be edited for each document.
\paragraph{Context Len}
Context Len refers to the context length that the model needs to process throughout the entire task, which is the length of the concatenated input and output.

\section{Dataset Samples and Detail Statistics}
\label{app:dataset_sample}
\subsection{Dataset Samples}
Figure \ref{fig:example_single} gives an example from \benchmarkname, where "Input" and "Target" correspond to the document before and after the update, respectively. "Facts" represents the facts to be edited, "Title" denotes the title of the document, which is also used as the subject during the update process, "Inputs\_sro", "Targets\_sro", and "Facts\_sro" represent the triples extracted from the input, output, and facts, respectively.

\begin{table*}[]
\centering
\begin{tcolorbox}[colback=gray!10,
			colframe=black,
			width=14cm,
			arc=2mm, auto outer arc,
			title={Data Example},	]

Title: Asagi Quscular

\vspace{0.5cm}
Inputs:
Ashaghy Gushchular () \textcolor{red}{or Ghushchular ()} is a village in the Shusha District of Azerbaijan. Until 2023 it was \textcolor{DarkYellow1}{controled} by the self-proclaimed Republic of Artsakh. The village had an Azerbaijani-majority population before the First Nagorno-Karabakh War. During the capture of the village, the Azerbaijani population was expelled, and it was reported that 8 civilians were killed.

\vspace{0.5cm}
Targets:
Ashaghy Gushchular () is a village in the Khojaly District of Azerbaijan. Until 2023 it was \textcolor{DarkYellow1}{controlled} by the self-proclaimed Republic of Artsakh. The village had an Azerbaijani-majority population before the First Nagorno-Karabakh War. During the capture of the village, the Azerbaijani population was expelled, and it was reported that 8 civilians were killed. \textcolor{deepgreen}{It was part of Shusha District with Malibeyli and Yuxari Quscular villages till 5 December 2023.}

\vspace{0.5cm}
Facts:
['Yuxari Quscular INTRODUCTION It was part of Shusha District with Malibeyli and Ashaghy Gushchular villages till 5 December 2023.', 'Malibeyli INTRODUCTION It was part of Shusha District with Asagi Quscular and Yuxari Quscular villages till 5 December 2023.']

\vspace{0.5cm}
Inputs\_sro:
[['Ashaghy Gushchular', 'instance of', 'village'], ['Shusha District', 'country', 'Azerbaijan'], ['Ashaghy Gushchular', 'ethnic group', 'Azerbaijani']]

\vspace{0.5cm}
Targets\_sro:
[['Ashaghy Gushchular', 'instance of', 'village'], ['Shusha District', 'country', 'Azerbaijan'], ['Ashaghy Gushchular', 'ethnic group', 'Azerbaijani'], \textcolor{deepgreen}{['Malibeyli', 'part of', 'Shusha District'], ['Yuxari Quscular', 'part of', 'Shusha District']}]

\vspace{0.5cm}
Facts\_sro:
[['Malibeyli', 'part of', 'Shusha District'], ['Yuxari Quscular', 'part of', 'Shusha District']]

\end{tcolorbox}	
    \caption{An example of \benchmarkname. In the input and target, \textcolor{red}{red} represents the parts deleted in the target, \textcolor{DarkYellow1}{yellow} represents the parts that have changed in the target, and \textcolor{deepgreen}{green} represents the newly added parts in the target. }
    \label{fig:example_single}
\end{table*}
\subsection{Data distribution}
\begin{figure}[!htbp]
\centering
\begin{tikzpicture}[scale=0.75]
\begin{axis}[
    ybar interval,
    xmin=115,
    xmax=5081,
    ymin=0,
    enlarge x limits=false,
    xtick={0,100,500,1000,1500,2000,2500,3000,3500,4000,4500,5000}, 
    xlabel style={at={(0.5,-0.1)}},
    extra x tick style={x tick label style={rotate=45, anchor=east}},
     xticklabel style={rotate=45},
    xlabel={Context Len},
    ylabel={Count},
    bar width=80, 
    xtick pos=bottom,    
    ytick pos=left,      
    axis lines=box,      
]
\addplot+[draw=black, fill={rgb,255:red,178;green,223;blue,138 } ] coordinates {
(363.3, 1912)
(611.6, 5170)
(859.9, 5827)
(1108.2, 4871)
(1356.5, 3495)
(1604.8, 2797)
(1853.1, 2162)
(2101.4, 2029)
(2349.7, 1945)
(2598, 1387)
(2846.3, 1095)
(3094.6, 1067)
(3342.9, 838)
(3591.2, 835)
(3839.5, 765)
(4087.8, 629)
(4336.1, 559)
(4584.4, 347)
(4832.7, 240)
(5081, 20)
};

\end{axis}
\end{tikzpicture}
\caption{\label{fig:dis_context}The distribution of Context length.}
\end{figure}
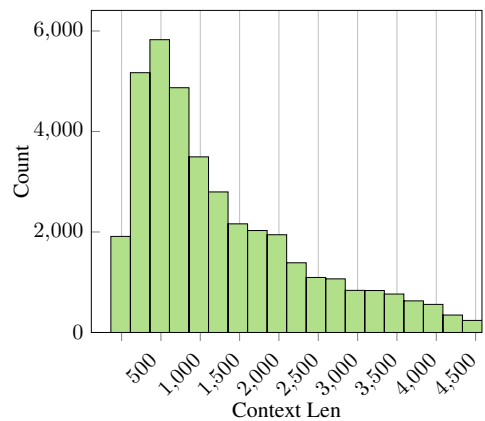

Table \ref{tab:detail_sta} presents the distribution of the length of facts and contexts used for each data item in \benchmarkname. Figure \ref{fig:dis_context} shows the distribution of context length. Figure \ref{fig:dis_fact} shows the distribution of fact length.

\begin{table}[!h]
    \centering


\begin{tabular*}{\hsize}{@{}@{\extracolsep{\fill}}ccccc@{}}
\midrule[2pt]
\multicolumn{5}{c}{Facts Per Data Item}                      \\
\midrule
1        & 2           & 3           & 4           & 5+      \\
18002        & 7233           & 3451           & 2078           & 7226       \\
\midrule[2pt]

\end{tabular*}


\begin{tabular*}{\hsize}{@{}@{\extracolsep{\fill}}cccc@{}}

\multicolumn{4}{c}{Context Len}                              \\
\midrule
0-512     & 512-2048    & 2048-4096 &   4096+ \\
4572        & 23283 & 8981           &1154       \\
  
\toprule[2pt]
\end{tabular*}
    \caption{Distribution of the Length of Facts and Contexts.}
    \label{tab:detail_sta}
\end{table}

\pgfplotsset{scaled y ticks=false}
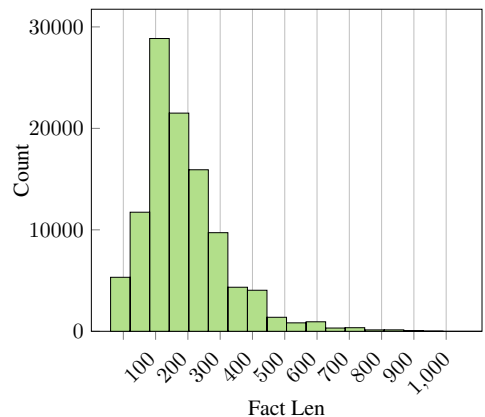
\begin{figure}[!htbp]
\centering
\begin{tikzpicture}[scale=0.75]
\begin{axis}[
    ybar interval,
    xmin=1,
    xmax=1211,
    ymin=0,
    enlarge x limits=false,
    xtick={100,200,300,400,500,600,700,800,900,1000,1100}, 
    xticklabel style={rotate=45},
    xlabel style={at={(0.5,-0.1)}},
    xlabel={Fact Len},
    ylabel={Count},
    bar width=50, 
    xtick pos=bottom,
    ytick pos=left,
    axis lines=box,
    yticklabel style={/pgf/number format/1000 sep=}
]

\addplot+[draw=black, fill={rgb,255:red,178;green,223;blue,138 }] coordinates {
(60.55, 5323)
(121.1, 11737)
(181.65, 28856)
(242.2, 21504)
(302.75, 15921)
(363.3, 9720)
(423.85, 4343)
(484.4, 4046)
(544.95, 1382)
(605.5, 833)
(666.05, 938)
(726.6, 323)
(787.15, 355)
(847.7, 139)
(908.25, 142)
(968.8, 57)
(1029.35, 16)
(1089.9, 0)
(1150.45, 0)
(1211, 11)
};

\end{axis}
\end{tikzpicture}
\caption{\label{fig:dis_fact}The distribution of fact length.}
\end{figure}

\section{Experimental Details}
\label{app:exp_detail}
\subsection{Experimental Environment}
All experiments were conducted on 2 $*$ NVIDIA A100-PCIE-40GB GPUs. The main software environment we used includes CUDA version 11.4, PyTorch version 2.0.1 \citep{Ansel_PyTorch_2_Faster_2024}, and Transformers library version 4.45.2 \citep{wolf-etal-2020-transformers}.

\subsection{Data}
We used all the data from \benchmarkname~ in our experiments. For each document edit, in the parameter modification methods (FT, MEMIT), we utilized all the editable facts, while in the RAG-based methods (IKE, SKEME, EREN), we used the top 5 retrieved facts to prevent exceeding the model's context length limit.

\subsection{Input Format}
To ensure that the large language model can correctly perform the document-level model editing task, we controlled the output format of the model. Specifically, we provided the original document to the model and instructed it to make updates. For all models, we used 1-shot learning to control the format while avoiding exceeding the model's context length limit. For RAG-based models (IKE, SKEME, EREN), we also prompted them to use the provided facts following previous research \citep{zeng2024fame,zheng2023can}. For EREN, we slightly modified the prompt of the original document to better align with our task. The complete list of prompts can be found in Appendix \ref{app:prompt}.

\subsection{Metrics Definition}
\label{app:Metrics format def}
In this section, we formally define the calculation methods for our metrics. 

First, let the original document be denoted as \( y \), the target document as \( y' \), and the facts to be edited as \( F = \{f_1, f_2, \dots, f_n\} \). The model's actual output is denoted as \( y'' \). Define the operation \( \text{Update}(a, b) \) as the computation of the difference between \( a \) and \( b \), and \( \text{Res}(a, b) \) as the part of \( a \) and \( b \) excluding the updated portions, i.e., the unchanged parts, \( E(x) \) as the extraction of triples from \( x \), and \( \text{ROUGE}(a, b) \) as the calculation of the ROUGE score between \( a \) and \( b \).

\paragraph{DR}
DR (Document ROUGE) represents the document update metric computed using ROUGE. Following \citet{logan2021fruit}, we actually consider only the updated sentences rather than the full texts, i.e.,
\[
DR = \text{ROUGE}(\text{Update}(y, y''), \text{Update}(y, y'))
\]

\paragraph{DE}
DE (Document Entity) is a document update metric calculated using the entity. Compared to ROUGE, it focuses more on the actual factual updates. We similarly compute only the updated entities, i.e.,
\[
\frac{\text{Update}(E(y), E(y'')) \cap \text{Update}(E(y), E(y')))}{\text{len}(\text{Update}(E(y), E(y')))}
\]
\paragraph{ER}
ER (Edit ROUGE) represents the ROUGE-based metric for triple updates. It is a finer-grained evaluation metric designed to assess whether each fact has been successfully updated. It is calculated as:
\[
ER = \text{ROUGE}(\text{Update}(y'', y), f_i)
\]
\paragraph{EE}
EE (Edit Entity) is the edit success rate calculated using triples. It is similarly used to evaluate whether each fact has been successfully updated. We calculate the proportion of updates in \( y'' \) relative to \( y \) that are supported by fact \( f_i \) in the updates of \( y' \) relative to \( y \), i.e.,

\[
A = \text{Update}(E(y), E(y''))
\]
\[
B_i = E(f_i) \cap \text{Update}(E(y), E(y'))
\]
\[
EE = \frac{A \cap B_i}{\text{len}(B_i)}
\]

\paragraph{RSE}
ROUGE Side Effect (RSE) is used to calculate the ROUGE score for the correctly retained parts of the document. It is computed based on the unmodified portions in the model's output and the expected output. Specifically, it is calculated as:

\[
RSE = \text{ROUGE}(\text{Res}(y, y''), \text{Res}(y, y'))
\]

\paragraph{ESE}

ESE (Entity Side Effect) is used to calculate the correctly retained entities in the document, which refers to the entities that remain unchanged in the model's output, similar to how ROUGE Side Effect (RSE) deals with the unmodified text. It is calculated as:

\[
\frac{\text{Res}(E(y), E(y'')) \cap \text{Res}(E(y), E(y')))}{\text{len}(\text{Res}(E(y), E(y')))}
\]

\subsection{Semantic Coherence Evaluation Criteria}
\label{app:sc}
Figure~\ref{fig:doc_eval_gen} shows the scoring criteria used to evaluate semantic coherence.

\begin{table*}[h!]
\centering
\begin{tcolorbox}[colback=gray!10, 
			colframe=black, 
			width=14cm, 
			arc=2mm, auto outer arc,
			title={Scoring Criteria},	]
0: The text is severely incoherent, with disjointed or contradictory ideas that lack logical flow. Sentences may be unrelated or nonsensical.

1: The text has partial coherence but contains noticeable inconsistencies, abrupt topic shifts, or repetitive/redundant statements that disrupt understanding.

2: The text is fully coherent, with clear logical progression, consistent ideas, and smooth transitions between sentences.
\end{tcolorbox}	
    \caption{Evaluation criteria for semantic coherence.}
    \label{fig:doc_eval_gen}
\end{table*}

\subsection{Implementation Details of Baselines}
\label{app:Implementation details}

For FT, MEMIT, and IKE, we use the framework provided by \citet{wang2023easyedit}\footnote{\url{https://github.com/zjunlp/EasyEdit}}. For EREN, we used the original implementation but modified the prompt to fit tasks\footnote{\url{https://github.com/thunlp/EREN}}.  For all methods, we use greedy decoding to obtain the LLM's output after editing and then perform evaluation.

\paragraph{FT}
Following previous research \citep{meng2022mass}, We apply Fine-Tuning (FT) to the given layer of the model. For GPT2-XL, we select layer 0, and for GPT-J and Llama2, we choose layer 21. 

\paragraph{MEMIT}
For GPT2-XL and GPT-J, we employ default hyperparameters. For Llama2, we update the parameters of layers $\{4, 5, 6, 7, 8\}$. Across all models, we calculate covariance statistics using 50,000 instances from Wikitext. We use the document title as the subject of the edited facts.

\paragraph{IKE}
We use the retrieval model settings from the original paper\footnote{\url{https://huggingface.co/sentence-transformers/all-MiniLM-L6-v2}}, retrieving the top 5 facts to prevent the output length from exceeding the LLM's context length limit.

\paragraph{EREN}
We adopt the retrieval model setup from the original paper\footnote{\url{https://huggingface.co/facebook/contriever}}, retrieving the top 5 facts to prevent the output length from exceeding the LLM limit. We slightly modified the prompt to adapt it to our task, which can be found in Section \ref{app:prompt}.

\paragraph{SKEME}
Due to the length of the document context, we use the title of the document as the subject in place of the subject generated by the LLM in the original paper to improve accuracy. We retain the top 5 facts to prevent the output length from exceeding the LLM limit.

\subsection{Prompts Used}
\label{app:prompt}
In this section, we present the prompts used for various methods. Figure \ref{fig:prompt_base} shows the prompts used for the base model, FT, and MEMIT. Figure \ref{fig:prompt_rag} displays the prompts used for IKE and SKEME, while Figure \ref{fig:prompt_eren} presents the prompts used for EREN.

\begin{figure*}[!t]
\centering
\begin{minipage}{\textwidth}
\begin{lstlisting}[]
Document: <few-shot input>
Updated version: <few-shot output>

Document: <input>
Updated version:
\end{lstlisting}
    \caption{Prompt for w/o Edit, FT and MEMIT. "Few-shot input" and "few-shot output" refer to the input and output of few-shot samples. The "input" contains the actual document to be updated.}
    \label{fig:prompt_base}
\end{minipage}
\end{figure*}

\begin{figure*}[!t]
\centering
\begin{minipage}{\textwidth}
\begin{lstlisting}[]
Document: <few-shot input>
Facts: <few-shot facts>
Updated version: <few-shot output>

Document: <input>
Facts: <retrieved facts>
Updated version:
\end{lstlisting}
    \caption{Prompt for IKE and SKEME. "Few-shot input" refers to the input of a few-shot samples. "Few-shot facts" refers to the facts to be edited in few-shot samples, which are directly obtained from the dataset. "Few-shot output" refers to the output of few-shot samples. The "input" contains the actual document to be updated, while the "retrieved facts" are the facts obtained through the retrieval method.}
    \label{fig:prompt_rag}
\end{minipage}
\end{figure*}

\begin{figure*}[!t]
\centering
\begin{minipage}{\textwidth}
\begin{lstlisting}[]
Read facts and update the document. If the document is unupdatable, say 'unupdatable'.

Document: <few-shot input>
Facts: <few-shot facts>
Updated version: <few-shot output>

Document: <input>
Facts: <retrieved facts>
Updated version:
\end{lstlisting}
    \caption{Prompt for EREN. System instructions were added to the figure \ref{fig:prompt_rag} to guide the model in following the task.}
    \label{fig:prompt_eren}
\end{minipage}
\end{figure*}
\section{Retrieval Result}
\label{app:retrval_result}
For all RAG models, we follow the EREN \citep{chen2024robust} setup and use \( \text{topk} = 5 \) to ensure a fair comparison.

Table \ref{tab:model_performance} shows the results of data retrieval for several RAG methods. It can be observed that IKE and EREN, which retrieve data via vector databases, have poorer retrieval results, while SKEME, which retrieves facts through entity-based searches, achieves better results.

\section{Relations}
Table \ref{tab:relation_statistics} presents the top 40 relations by frequency, extracted from the triples in \benchmarkname.
\begin{table*}[!ht]
    \centering
    \begin{tabular}{cccc}
    \hline
    \textbf{Wikidata Label} &  \textbf{Wikidata ID} & \textbf{Relevant Triples} & \textbf{The proportion of all relations}  \\
    
    \hline
of & P642 & 243694 & 17.27 \% \\ 
including & P1012 & 31768 & 2.25 \% \\ 
competition won & P2522 & 30580 & 2.17 \% \\ 
replaced by & P1366 & 21881 & 1.55 \% \\ 
award received & P166 & 19690 & 1.4 \% \\ 
founded by & P112 & 18149 & 1.29 \% \\ 
destroyed & P3082 & 16459 & 1.17 \% \\ 
in operation on service & P10788 & 15441 & 1.09 \% \\ 
name & P2561 & 15143 & 1.07 \% \\ 
release of & P9831 & 15102 & 1.07 \% \\ 
had as last meal & P3902 & 15020 & 1.06 \% \\ 
position held & P39 & 14608 & 1.04 \% \\ 
has cause & P828 & 14524 & 1.03 \% \\ 
location & P276 & 13992 & 0.99 \% \\ 
has tense & P3103 & 13573 & 0.96 \% \\ 
has use & P366 & 13138 & 0.93 \% \\ 
start point & P1427 & 13059 & 0.93 \% \\ 
uses & P2283 & 12609 & 0.89 \% \\ 
followed by & P156 & 10576 & 0.75 \% \\ 
lighting & P8228 & 9286 & 0.66 \% \\ 
acknowledged & P7137 & 8804 & 0.62 \% \\ 
maintained by & P126 & 8617 & 0.61 \% \\ 
contains & P4330 & 8376 & 0.59 \% \\ 
time played & P9140 & 7862 & 0.56 \% \\ 
develops from & P3094 & 7760 & 0.55 \% \\ 
moved by & P6939 & 7702 & 0.55 \% \\ 
named after & P138 & 7688 & 0.54 \% \\ 
place of birth & P19 & 7542 & 0.53 \% \\ 
location of formation & P740 & 7334 & 0.52 \% \\ 
points for & P1358 & 7032 & 0.5 \% \\ 
follows & P155 & 6919 & 0.49 \% \\ 
made from material & P186 & 6852 & 0.49 \% \\ 
for work & P1686 & 6716 & 0.48 \% \\ 
dedicated to & P825 & 6552 & 0.46 \% \\ 
produced by & P2849 & 6502 & 0.46 \% \\ 
merged into & P7888 & 6489 & 0.46 \% \\ 
towards & P5051 & 6485 & 0.46 \% \\ 
damaged & P3081 & 6384 & 0.45 \% \\ 
based on & P144 & 6271 & 0.44 \% \\ 
winner & P1346 & 6227 & 0.44 \% \\ 
 
    \hline
    \end{tabular}
    \caption{Statistics of relations. Wikidata Label refers to the name of the relation in Wikidata, while Wikidata ID refers to the ID assigned to it by Wikidata.}
    \label{tab:relation_statistics}
\end{table*}

\begin{table}[!ht]
\centering
\begin{tabular}{cccc}
\hline
\textbf{Method} & \textbf{Precision} & \textbf{Recall} & \textbf{F1 Score} \\ \hline
IKE & 47.20 & 89.71 & 53.45 \\ 
EREN & 47.60 & 89.86 & 53.77 \\ 
SKEME & 100.00 & 93.03 & 95.55 \\ \hline
\end{tabular}

\caption{Retrieval Result}
\label{tab:model_performance}
\end{table}

\section{Running Example}
Table \ref{tab:running_example} presents a running example obtained on Llama2.
\begin{table*}[!ht]
\centering
\begin{tabular}{c|c|p{12cm}}

    \midrule[1.5pt]

\multirow{3}*{Data}   & Input    & The Belle Bay Formation is a metamorphosed formation cropping out in Newfoundland. \\
    \cline{2-3}
                     ~   & Target   & The Belle Bay Formation is a metamorphosed formation cropping out in Newfoundland. It is locally overlain by of the Andersons Cove Formation. \\
    \cline{2-3}
                   ~       & Facts    & Andersons\_Cove\_Formation INTRODUCTION The Andersons Cove Formation is a formation which overlies the Belle Bay Formation in Newfoundland, Canada. \\
    \midrule[1.5pt]
\multirow{5}*{Output} & w/o Edit & The Belle Bay Formation is a metamorphosed formation cropping out in Newfoundland. It is exposed in high-grade gneiss terranes east of Belle Bay. It is metamorphosed at a contact with the Bonavista formation. It crops out in a similar high-pressure setting in eastern Quebec (O'Hare, 1987, Geological Survey of Canada, p. 42) and near Gogama, Ontario (Lorna, 1979; Ridgway, 1985). \\
    \cline{2-3}
                    ~      & FT       &  The Belle Bay Formation is a metamorphosed formation cropping out in Newfoundland, Canada. It is a member of the North America-wide group of rock formations known as the Taconian Orogeny. The Belle Bay Formation is over 800 million years old.
 \\
    \cline{2-3}
                     ~     & MEMIT    &   The Bay Formation, a late Devonian geologic formation, is exposed along Southeastern Newfoundland between Bay Bulls, Bay de Verde, L'Anse aux Meadows, and Point Amour on the Northern Peninsula. \\
    \cline{2-3}
                     ~     & IKE      &  The Belle Bay Formation is a metamorphosed formation cropping out in Newfoundland, Canada. It is overlain by the Andersons Cove Formation. \\
    \cline{2-3}
                     ~     & SKEME    & The Belle Bay Formation is a metamorphosed formation cropping out in Newfoundland, Canada. It is overlain by the Andersons Cove Formation. \\
    \midrule[1.5pt]
\end{tabular}
\caption{The running example obtained on Llama2, where "data" refers to the input data, and "output" refers to the results generated by each method. Note that, since this data item is relatively simple, the facts retrieved by IKE and SKEME are the same, so the outputs do not differ.\label{tab:running_example}}
\end{table*}

\end{document}